\documentclass[letterpaper]{article} 
\usepackage{aaai24}  
\usepackage{times}  
\usepackage{helvet}  
\usepackage{courier}  
\usepackage[hyphens]{url}  
\usepackage{graphicx} 
\usepackage{natbib}  
\usepackage{caption} 
\usepackage{algorithm}
\usepackage{algorithmic}

\usepackage{listings}
\usepackage{color}

\usepackage{amstext}
\usepackage{amsmath}
\usepackage{amssymb}
\usepackage{multirow}
\usepackage{booktabs}
\usepackage{array}
\usepackage{subcaption}
\usepackage{natbib}
\usepackage{cite}

\newcommand{\tablestyle}[2]{\setlength{\tabcolsep}{#1}\renewcommand{\arraystretch}
{#2}\centering\footnotesize}

\newtheorem{definition}{Definition}

\DeclareCaptionStyle{ruled}{labelfont=normalfont,labelsep=colon,strut=off} 
\floatstyle{ruled}
\newfloat{listing}{tb}{lst}{}
\floatname{listing}{Listing}
\title{Decoupled Contrastive Multi-View Clustering with High-Order Random Walks}
\author{
    Yiding Lu, Yijie Lin, Mouxing Yang, Dezhong Peng, Peng Hu\textsuperscript{\rm*}, Xi Peng\thanks{Corresponding authors}
}
\affiliations{
    College of Computer Science, Sichuan University\\
    \{yidinglu.gm, linyijie.gm, yangmouxing, penghu.ml, pengx.gm\}@gmail.com,~pengdz@scu.edu.cn
}

\begin{document}

\maketitle

\begin{abstract}
In recent, some robust contrastive multi-view clustering (MvC) methods have been proposed, which construct data pairs from neighborhoods to alleviate the false negative issue, \textit{i.e.}, some intra-cluster samples are wrongly treated as negative pairs. Although promising performance has been achieved by these methods, the false negative issue is still far from addressed and the false positive issue emerges because all in- and out-of-neighborhood samples are simply treated as positive and negative, respectively. To address the issues, we propose a novel robust method, dubbed decoupled contrastive multi-view clustering with high-order random walks (DIVIDE). In brief, DIVIDE leverages random walks to progressively identify data pairs in a global instead of local manner. As a result, DIVIDE could identify in-neighborhood negatives and out-of-neighborhood positives. Moreover, DIVIDE embraces a novel MvC architecture to perform inter- and intra-view contrastive learning in different embedding spaces, thus boosting clustering performance and embracing the robustness against missing views. To verify the efficacy of DIVIDE, we carry out extensive experiments on four benchmark datasets comparing with nine state-of-the-art MvC methods in both complete and incomplete MvC settings. The code is released on \url{https://github.com/XLearning-SCU/2024-AAAI-DIVIDE}.
\end{abstract}

\begin{figure}[t]
    \centering
\includegraphics[width=\columnwidth]{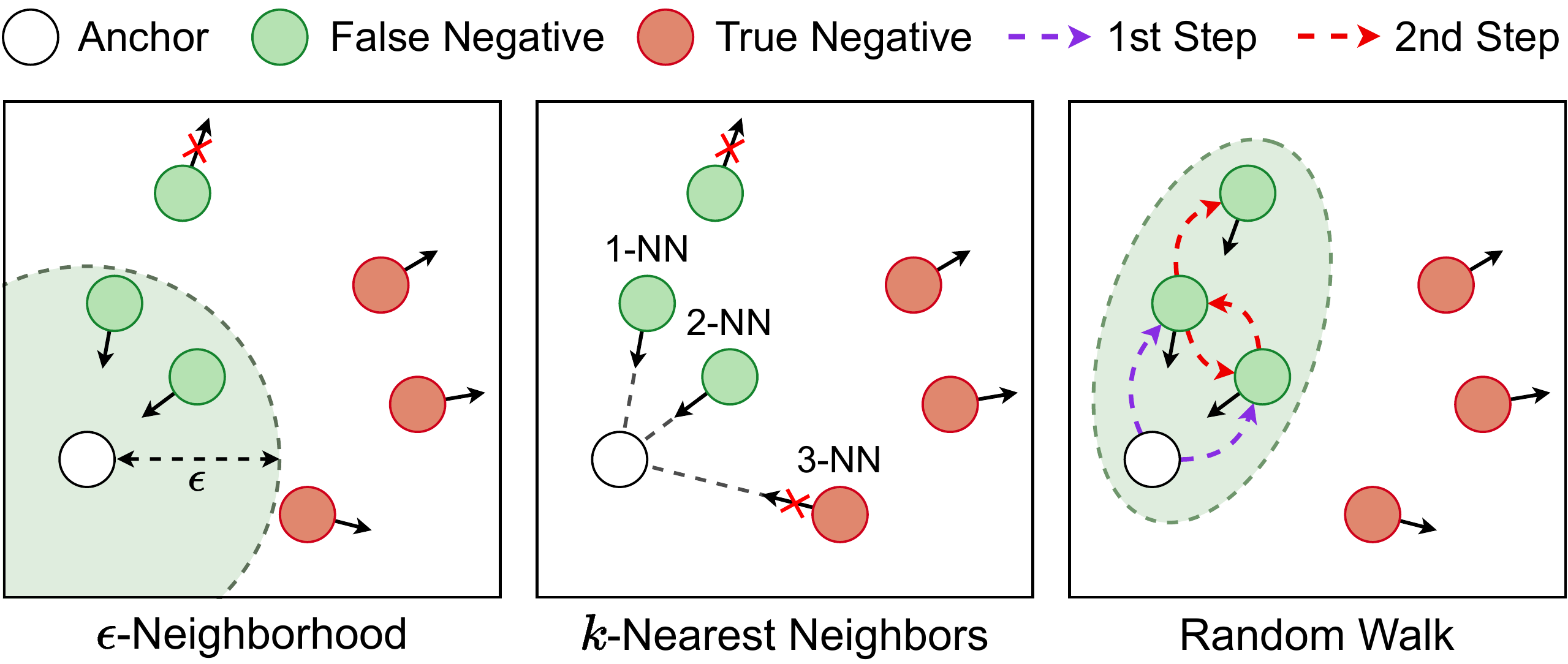}
    \caption{Our observation and basic idea. As illustrated, the existing robust contrastive MvC methods construct negative pairs resorting to $\epsilon$- or $k$-nearest-neighborhood approaches, which will treat all in-neighborhood samples as positive and out-of-neighborhood samples as negative, resulting in the false positive and false negative issues. Different from these methods, our method identifies the data pairs in a global instead of local manner by performing multi-step random walks on an affinity graph. Thanks to the globality, on the one hand, our method will treat some in-neighborhood samples as negative if high-order neighbors are with a lower affinity score, thus avoiding the construction of FPs to some extent. On the other hand, our method will treat some out-of-neighborhood samples as positive, thus avoiding the construction of FNs in part.}
\label{fig:observation}
\end{figure}

\section{Introduction}
Multi-view data, which can capture diverse and complementary aspects of the same object from different sources or modalities~\cite{DM2C,lin2022graph}, has become increasingly prevalent in many real-world applications such as human activity recognition~\cite{HAR2021review} and person re-identification~\cite{yang2022DART}. As one of the most effective tools to analyze multi-view data, multi-view clustering~(MvC) could group unlabeled instances into semantic clusters by exploring and exploiting the latent information hidden in different views. In other words, multi-view representation learning lies at the heart of MvC~\cite{BMVC,CPMNets-jour}.

As a powerful unsupervised representation learning method, contrastive learning~\cite{MoCo2020,simclr}, which aims to compact semantically similar pairs (\textit{e.g.}, different views of the same instance) and scatter dissimilar pairs (\textit{e.g.}, different views of different instances) in the embedding space, has achieved state of the arts in MvC~\cite{MVSCN,CPMNets,CoMVC}. Despite the success achieved by the contrastive MvC methods, they have suffered from the false negative (FN) issue~\cite{yang2022SURE}. To be specific, most existing contrastive MvC methods will randomly choose some samples to construct negative pairs, and thus some intra-cluster samples are wrongly treated as negative pairs with a high probability. In brief, suppose the data could be grouped into $C$ clusters, there is a probability of $1/C$ to select the intra-cluster samples as negative. To alleviate the performance degradation caused by FNs, some studies~\cite{GCC2021,yang2022SURE,FNC_Boosting, zheng2022contrastive, sun2023learning} have proposed to rectify FNs through $\epsilon$-neighborhood or $k$-nearest-neighbors. In brief, for each anchor, its $\epsilon$-neighborhood or $k$-nearest-neighbors are treated as positive and the remainder is treated as negative. Although the methods have shown remarkable effectiveness in the FNs correction, they will suffer from under- and over-rectification problems. To be exact, the out-of-neighborhood intra-cluster samples will still be treated as negatives, and the in-neighborhood inter-cluster samples might be treated as positives. Notably, in the following, the out-of- and in-neighborhood refers to the first-order relation only. 

To address the false negative and false positive (FP) challenges caused by the under- and over-rectification problems, we propose Decoupled contrastIve multi-View clusterIng with high-orDEr random walk (DIVIDE). As illustrated in Fig.~\ref{fig:observation}, given an affinity graph calculated in the embedding space, DIVIDE first performs a random walk for each anchor by choosing a neighbor to move according to the affinity. 
By applying multi-step random walks, we progressively obtain the affinity between the anchor and the high-order neighbors, and thus the out-of-neighborhood intra-cluster samples could be identified and accordingly the FN issue is alleviated. As another merit of the high-order affinity, the in-neighborhood inter-cluster samples could also be identified and the FP issue is addressed accordingly because the first-order neighbors might be with a smaller high-order affinity. Besides the above contributions, we propose a novel MvC architecture that performs intra-view and inter-view contrastive learning in different spaces with the help of a cross-view decoder. Thanks to the decoupled architecture, the view-specific information could be preserved while learning the cross-view consistency, thus improving MvC performance. 
In brief, the main contributions of this work could be summarized as follows: 

\begin{itemize}
    \item To alleviate the under- and over-rectification problems faced by existing contrastive MvC methods, we propose using a random walk approach to obtain the high-order affinity so that the in-neighborhood FPs and out-of-neighborhood FNs could be discovered. 
    \item We design a novel decoupled contrastive MvC framework. Different from existing methods, our framework performs intra- and inter-view contrastive learning in different instead of the same space, thus being in favor of the intra-view information and cross-view consistency. 
    \item Extensive experiments verify the superiority of DIVIDE over nine methods on four benchmarks in complete and incomplete multi-view settings. 
\end{itemize}

\begin{figure*}[t]
    \centering
    \includegraphics[width=0.96\textwidth]{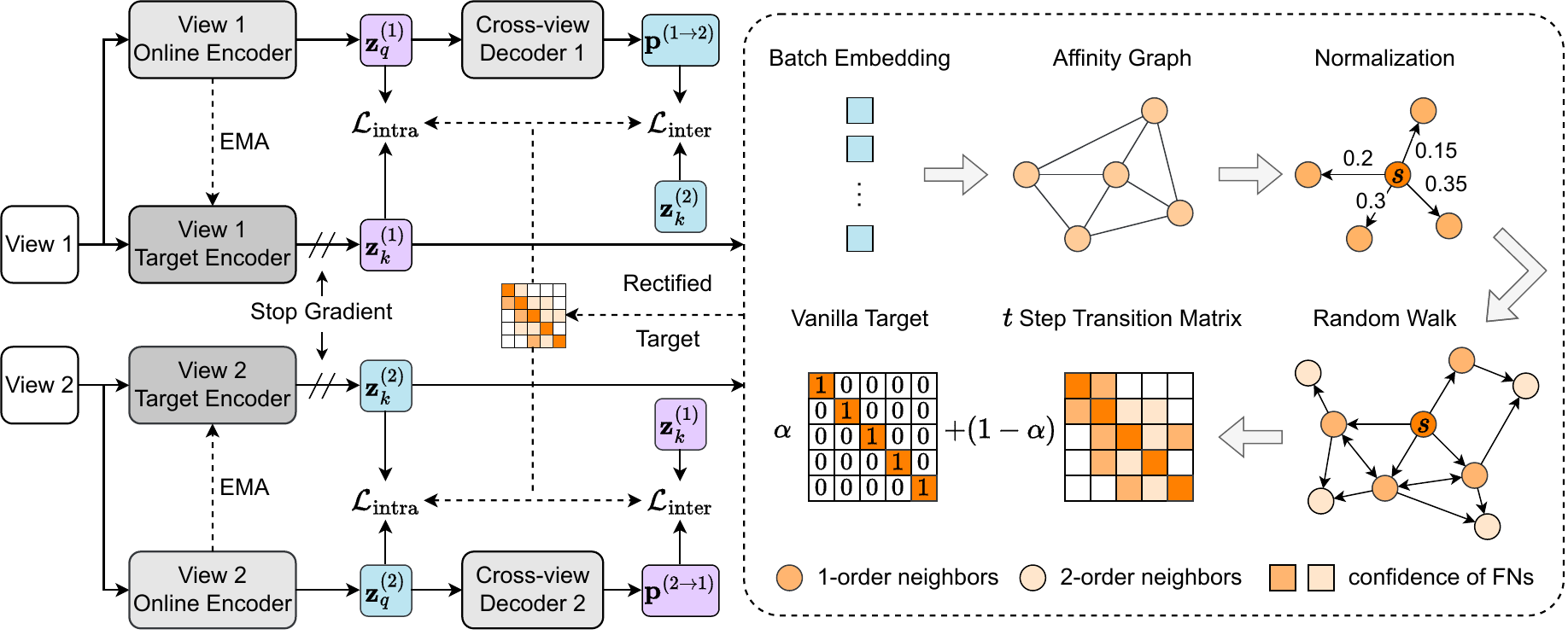}
    \caption{Overview of DIVIDE. 
    For clarity, we use $V=2$ as an example for illustration. 
    DIVIDE consists of two modules: decoupled contrastive learning framework (the left part) and random-walk-based FNs correction (the right part). Left: data of each view is first passed into the view-specific Siamese encoders to obtain the dual-embedding $\mathbf{z}^{(v)}_{q}$ and $\mathbf{z}^{(v)}_{k}$. 
    After that, the cross-view decoder projects $\mathbf{z}^{(v)}_{q}$ into a new subspace.
    Finally, the decoupled contrastive losses are applied in two subspaces to learn cross-view consistency and preserve view-specific information; Right: The affinity graph is constructed from each data batch and the corresponding adjacency matrix is normalized as the initial transition matrix.
    After that, DIVIDE performs a random walk for each anchor by choosing a neighbor to move according to the affinity. By adopting $t$-step random walks for all anchors, DIVIDE progressively identifies the corresponding high-order neighbors (\textit{i.e.}, false negatives) with a probability matrix which is formulated as a $t$-step transition matrix $\mathbf{M}^{t}$.
    Finally, DIVIDE uses $\mathbf{M}^{t}$ to rectify the target of decoupled contrastive losses so that the false negative issue can be addressed.}
    \label{fig:Overview}
\end{figure*}

\section{Related Work} 

In this section, we briefly review recent developments in deep MvC and contrastive MvC.

\subsection{Deep Multi-view Clustering}

From a self-supervised perspective, most existing deep multi-view clustering methods can be divided into two categories: reconstruction- and contrast-based methods. To be specific, the former typically focuses on preserving intra-view information (\textit{i.e.,} complementarity) by using multiple autoencoders~\cite{CPMNets,CPMNets-jour,AE2nets} to extract compact representations. As the reconstruction loss might retain clustering-agnostic redundancy-like backgrounds that are further used for reconstruction~\cite{COMPLETER}, the MvC performance would suffer from degradation. 
In a different way, the contrast-based methods~\cite{DCP2022,CoMVC,trosten2023effects} mainly attempt to explore consistent information across different views, which could preserve the discrimination while removing the redundancy from representations. Although contrast-based methods have achieved promising performance, the learned representations might collapse into nearly identical ones~\cite{CrossCLR}, hence losing view-specific information. 

Different from these existing approaches, our decoupled contrastive framework could preserve view-specific information through cross-view reconstruction, thus enjoying the merits of both the reconstruction- and contrast-based MvC methods. Furthermore, our framework can naturally tackle the missing views thanks to the cross-view reconstruction paradigm. In other words, it does not require elaborate extensions for incomplete settings like most existing MvC methods~\cite{xu2019adversarial,diffusion_incomplete2023}.

\subsection{Contrastive Multi-view Clustering}
Like the vanilla contrastive learning~\cite{MoCo2020,simclr,liu2022scc}, contrastive MvC aims to align representations of the data from different views by maximizing the similarity between positive pairs and minimizing that of negative pairs, where the known cross-view samples are treated as positives and the others are treated as negative. Clearly, such a data pair construction method will result in the false negative issue because massive intra-cluster samples are treated as negatives. To address the FN issue, several works have been conducted. For example, \citet{DebiasedCL} proposed modeling an unbiased negative probability distribution to reweight the negative terms on the denominator of InfoNCE so that the influence of FNs is alleviated.  \citet{GCC2021,FNC_Boosting} proposed using $k$-nearest-neighbors with the given sample to form the positive pairs, thus reducing the intra-cluster FNs. Moreover, 
\citet{yang2022SURE} proposed dividing positive and negative samples through thresholding a provable similarity metric. In other words, it implicitly constructs positive pairs from a $\epsilon$-neighborhood. 

Although these methods have achieved robustness against FNs, most of them have suffered from under- and over-rectification problems as elaborated in Introduction. To solve the problems, unlike these robust methods that construct data pairs from the first-order neighborhood, our method constructs positive pairs from the high-order neighborhoods. More specifically, our DIVIDE employs a random walk approach to discover the out-of-neighborhood positives, thus alleviating the false negative issue. As another merit of the high-order neighborhood, our method will not treat the in-neighborhood negatives as positives, namely, alleviating the false positive issue. 

\section{Method}
In this section, we elaborate on the proposed method, \textit{i.e.}, Decoupled contrastIve multi-View clusterIng with high-orDEr random walks (DIVIDE). As shown in Fig.~\ref{fig:Overview}, DIVIDE consists of two modules, namely, i) a decoupled contrastive learning framework that is designed to learn cross-view consistency while preserving view-specific information, and ii) a random-walk-based rectifier that is proposed to detect and correct FNs.

\subsection{Decoupled Contrastive Learning Framework}
Given a multi-view dataset $\{(\mathbf{x}^{(1)}_i, \dots, \mathbf{x}^{(V)}_i)\}_{i=1}^N$ with $N$ instances observed from $V$ views, contrastive MvC methods~\cite{ProImp2023,CoMVC,DCP2022} aim at learning a common representation shared by different views through maximizing the consistency between the embedding extracted from different view-specific encoders. Due to overemphasizing cross-view consistency learning, the view-specific information would be discarded~\cite{tsai2020self, wang2022rethinking}, resulting in performance degradation~\cite{yang2022SURE, CPMNets-jour}. To overcome this challenge, we design a novel MvC framework that consists of view-specific Siamese encoders, cross-view decoders, and decoupled contrastive loss.

\paragraph{View-specific Siamese Encoders.}
For each view, the Siamese encoder consists of an online encoder $f^{(v)}_{q}$ and a target encoder $f^{(v)}_{k}$ which is a momentum version~\cite{MoCo2020} of $f^{(v)}_{q}$. More specifically, $f^{(v)}_{k}$ shares the same architecture with $f^{(v)}_{q}$ and is updated using the Exponential Moving Average (EMA) of $f^{(v)}_{q}$. 

For a given mini-batch of instances, we first feed them into $f^{(v)}_{q}$ and $f^{(v)}_{k}$ to obtain the corresponding view-specific embedding, \textit{i.e.},

\begin{equation*}
\begin{aligned}
    &\mathbf{z}^{(v)}_{q, i}=f^{(v)}_{q}(\mathbf{x}^{(v)}_{i}),\\
    &\mathbf{z}^{(v)}_{k, i}=f^{(v)}_{k}(\mathbf{x}^{(v)}_{i}).
\end{aligned}
\end{equation*}

\paragraph{Cross-view Decoders.}
As mentioned earlier, simply maximizing the consistency between cross-view embedding may result in the loss of view-specific information. As a remedy, we design a novel cross-view decoder that could preserve the view-specific information without overemphasizing the cross-view consistency. To be specific, we utilize the cross-view decoder $g^{(v\rightarrow u)}$ to project the embedding $\mathbf{z}^{(v)}_{q, i}$ into the embedding space of another view $u$, and output the embedding $\mathbf{p}^{(v\rightarrow u)}_{i}$, \textit{i.e.},

\begin{equation*}
    \mathbf{p}^{(v\rightarrow u)}_{i}=g^{(v\rightarrow u)}(\mathbf{z}_{q, i}^{(v)}).
\end{equation*}

Thanks to the cross-view decoders, our framework enjoys the following two merits. On the one hand, consistency learning is performed in two view-specific spaces instead of a single common space like existing methods~\cite{lin2022contrastive}. As a result, $\mathbf{z}_{q}^{(v)}$ would achieve the cross-view consistency while preserving the view-specific complementary information, which remarkably boosts the MvC performance as verified in Table~\ref{table:ablation_stop_gradient}.
On the other hand, the samples could be recovered across different views via the cross-view decoders, thus endowing our method with incomplete multi-view clustering~\cite{xu2019adversarial,DSIMVC} as shown in Table~\ref{table:main}.                                     
\paragraph{Decoupled Contrastive Loss.}
As elaborated above, the well-designed encoders and decoders aim to decouple the intra- and inter-view learning from a common space into two view-specific spaces. 
Accompanying the novel architecture, we propose the following decoupled contrastive loss:

\begin{equation}
    \resizebox{0.95\columnwidth}{!}{$
    \begin{aligned}
    \mathcal{L}&=\mathcal{L}_{\text{intra}} + \mathcal{L}_{\text{inter}} \\
    &=\underbrace{\sum_{v=1}^{V}\mathcal{H}(\mathbf{T}^{(v)},\rho(\mathbf{z}^{(v)}_{q},\mathbf{z}^{(v)}_{k}))}_{\text{Intra-view Contrastive}}+\underbrace{\sum_{v\neq u}^{V}\mathcal{H}(\mathbf{T}^{(u)},\rho(\mathbf{p}^{(v\rightarrow u)},\mathbf{z}^{(u)}_{k}))}_{\text{Inter-view Contrastive}},\\
    \end{aligned}
    $}
\label{eq:loss}
\end{equation}

where $\mathcal{H}(p,q)$ denotes the cross entropy, $\mathbf{T}\in\mathbb{R}^{n\times n}$ is the pseudo target (Eq.~\ref{eq:t}) that indicates the given pair is positive or negative, and $\rho(\mathbf{a},\mathbf{b})$ is the pairwise similarity $s(\cdot, \cdot)$ with the row-wise normalization operator, 
\textit{i.e.},
\begin{equation}
    \left[\rho(\mathbf{a},\mathbf{b})\right]_{ij}=\frac{\exp(s(\mathbf{a}_i,\mathbf{b}_j)/\tau)}{\sum_{l=1}^n\exp(s(\mathbf{a}_i,\mathbf{b}_l)/\tau)},
\end{equation}
where $\tau$ is the temperature fixed to 0.5 throughout our experiments. 

In general, most existing contrastive MvC methods~\cite{CoMVC, ProImp2023} usually use the off-the-shelf instances as positive pairs and construct negative pairs using random sampling, \textit{i.e.}, $\mathbf{T}$ is set as identity matrix $\mathbf{I}_n$.
However, as mentioned in Introduction, it would inevitably introduce some FNs, thus misleading the model optimization. 
In the following, we will elaborate on how to endow our method with robustness against FNs.

\subsection{False Negative Identification via Random Walks}
\label{sec:FN rectifier}
To achieve robustness against FNs, one feasible solution is identifying the potential FNs from negative pairs and rectifying the pseudo target accordingly. 
To this end, most existing robust contrastive MvC methods~\cite{GCC2021, yang2022SURE} enlarge the domain of positives and accordingly reduce the amounts of FNs resorting to $\epsilon$-neighborhood or $k$-nearest-neighbors. 

However, such a vanilla neighborhood-based paradigm may cause under- and over-rectification. Namely, the out-of-neighborhood samples within the same cluster would still be treated as negatives while the neighboring inter-cluster samples would be mistakenly treated as positive.
To reduce the amounts of FNs and avoid introducing FPs, we propose employing multi-step random walks to progressively identify the high-order neighbors for each anchor. 
With the high-order neighbors and the corresponding affinity, our method could model the probability of negative pairs being FNs. For ease of presentation, we briefly introduce the preliminary on the random walk. To be exact, a random walk on a graph is a process that starts at some nodes and moves to other nodes according to the edge weights at each step~\cite{rw_on_graph}. Formally, the random walk on a graph could be represented by a transition matrix defined as follows.  
\begin{definition}[Transition Matrix] 
For an undirected graph $\mathcal{G}$ with $n$ nodes, the random walk transition matrix on this graph is given by
\begin{equation}
    \mathbf{M}=\mathbf{AD}^{-1}
    \label{Markov_Normalize}
\end{equation}
where $\mathbf{A}$ is the adjacency matrix of this graph, $\mathbf{A}_{ij}$ denotes the edge weight, $\mathbf{D}= \operatorname{diag}(d_1,d_2,\dots,d_N)$, and $d_i=\sum_{j}\mathbf{A}_{ij}$.
\end{definition}

Clearly, $\mathbf{M}_{ij}$ herein denotes the probability of moving from the $i$-th node to the $j$-th node in one step. Similarly, the probability of the  position after multi-step random walks could be denoted by: 
\begin{equation}
    p(t)=p(t-1)\mathbf{M}=\cdots=p(0)\mathbf{M}^t
\end{equation}
where $\mathbf{M}^t$ is $t$-th power of the transition matrix $\mathbf{M}$ and $t$ is the transmission step.

To identify FNs from the negative pairs, we use each anchor as the starting node and perform $t$ step random walks to find the potential FNs from the high-order neighbors. More specifically,  we first construct a fully-connected affinity graph for the in-batch instances by regarding the embedding as nodes and defining the edge weights with the heat kernel similarity:
\begin{equation}
    \mathbf{A}^{(v)}_{ij}=\exp(-||\mathbf{z}^{(v)}_{k,i}-\mathbf{z}^{(v)}_{k,j}||^{2}/\sigma),
\end{equation}
where $\mathbf{z}^{(v)}_{k,i}$ is the $i$-th anchor embedding,  $\mathbf{z}^{(v)}_{k,j}$ is the corresponding negative embedding,
and $\sigma$ is a hyper-parameter fixed as 0.1 throughout our experiments. 

After that, one could obtain the random walk transition matrix $\mathbf{M}^{(v)}$ by normalizing the adjacency matrix $\mathbf{A}^{(v)}$ in a row-wise manner. 
Similarly, the multi-step transition matrix ${\mathbf{M}^{(v)}}^{t}$ could be obtained, where ${\mathbf{M}^{(v)}}^{t}_{ij}$ denotes the probability of the $j$-th negative  being the $t$-th order neighbor (\textit{i.e.}, FNs) for the $i$-th anchor.
In other words, the entry of ${\mathbf{M}^{(v)}}^{t}$ could serve as the soft target for the contrastive loss (Eq.~\ref{eq:loss}), which denotes the correlation probability of given two samples.

Finally, we use $\mathbf{M}^{t}$ as the pseudo target of Eq.~\ref{eq:loss} to achieve the robustness against FNs, \textit{i.e.},
\begin{equation}
    \mathbf{T}^{(v)}=\alpha \mathbf{I}_n+(1-\alpha){\mathbf{M}^{(v)}}^t
    \label{eq:t}
\end{equation}
where $\alpha$ is a balanced parameter fixed to 0.5 in our experiments. 
In the implementation, we construct the affinity graphs for each view and accordingly obtain $\mathbf{T}^{(v)}$.
For the intra-view contrastive learning of the $v$-th view, $\mathbf{T}^{(v)}$ is utilized as the target to improve intra-view discrimination. Such a self-rectified target could better preserve the view-specific information, thus benefitting MvC performance as evaluated in our experiments. In contrast, the affinity of the predicted view is used to construct cross-view target $\mathbf{T}^{(u)}$ for the inter-view contrastive learning. Such a swap-rectified target could improve the cross-view interactions, leading to better cross-view consistency.
For a more comprehensive understanding, we conduct ablation study on different target construction strategies in Table~\ref{table:compare_strategy}.

\begin{table*}[t]
    \tablestyle{4.5pt}{1.01}{
        \begin{tabular}{c|l|ccc|ccc|ccc|ccc}
            \toprule
            \multirow{2}{*}{Setting}     & \multicolumn{1}{c|}{\multirow{2}{*}{Method}} & \multicolumn{3}{c|}{Scene15}                  & \multicolumn{3}{c|}{Caltech101}               & \multicolumn{3}{c|}{Reuters}                  & \multicolumn{3}{c}{LandUse21}                \\ 
                                         & \multicolumn{1}{c|}{}                        & ACC           & NMI           & ARI           & ACC           & NMI           & ARI           & ACC           & NMI           & ARI           & ACC           & NMI           & ARI           \\ \midrule
            \multirow{10}{*}{Incomplete} & DCCAE\cite{DCCAE}                                        & 29.0          & 29.1          & 12.9          & 29.1          & 58.8          & 23.4          & 47.0          & 28.0          & 14.5          & 14.9          & 20.9          & 3.7           \\
                                         & BMVC\cite{BMVC}                                         & 32.5          & 30.9          & 11.6          & 40.0          & 58.5          & 10.2          & 32.1          & 7.0           & 2.9           & 18.8          & 18.7          & 3.7           \\
                                         & PMVC\cite{PMVC}                                         & 25.5          & 25.4          & 11.3          & 50.3    & 74.5    & 41.5          & 29.3          & 7.4           & 4.4           & 20.0          & 23.6          & 8.0           \\
                                         & DAIMC\cite{DAIMC}                                        & 27.0          & 23.5          & 10.6          & \underline{56.2}          & \underline{78.0}          & 41.8          & 40.9          & 18.7          & 15.0          & 19.3          & 19.5          & 5.8           \\
                                         & EERIMVC\cite{EERIMVC}                                      & 28.9          & 27.0          & 8.4           & 43.6          & 69.0          & 26.4          & 29.8          & 12.0          & 4.2           & 22.1          & 25.2          & 9.1           \\
                                         & SURE\cite{yang2022SURE}                                         & 39.6          & 41.6          & 23.5          & 34.6          & 57.8          & 19.9          & 47.2          & 30.9          & 23.3          & \underline{23.1}    & \underline{28.6}    & \underline{10.6}    \\
                                         & DSIMVC\cite{DSIMVC}                                       & 30.6          & 35.5          & 17.2          & 16.4          & 24.8          & 9.2           & 39.9          & 19.6          & 17.1          & 18.6          & 18.8          & 5.7           \\
                                         & DCP\cite{DCP2022}                                          & 39.5          & 42.4          & 23.5          & 44.3          & 71.0          & \underline{45.3}    & 34.6          & 17.5          & 2.9           & 22.2          & 27.0          & 10.4          \\
                                         & ProImp\cite{ProImp2023}                                       & \underline{41.6}    & \underline{42.9}    & \underline{25.3}    & 36.3          & 65.4          & 25.4          & \underline{51.9}    & \underline{35.5}    & \underline{28.5}    & 22.4          & 26.6          & 9.9           \\
                                         & \textbf{DIVIDE(Ours)}                                          & \textbf{46.8} & \textbf{45.7} & \textbf{29.1} & \textbf{63.4} & \textbf{82.5} & \textbf{52.4} & \textbf{54.7} & \textbf{37.3} & \textbf{28.6} & \textbf{30.0} & \textbf{35.8} & \textbf{16.0} \\ \midrule
            \multirow{10}{*}{Complete}   & DCCAE\cite{DCCAE}                                        & 34.6          & 39.0          & 19.7          & 45.8          & 68.6          & 37.7          & 42.0          & 20.3          & 8.5           & 15.6          & 24.4          & 4.4           \\
                                         & BMVC\cite{BMVC}                                         & 40.5          & 41.2          & 24.1          & 50.1          & 72.4          & 33.9          & 42.4          & 21.9          & 15.1          & 25.3          & 28.6          & 11.4          \\
                                         & PMVC\cite{PMVC}                                         & 30.8          & 31.1          & 15.0          & 40.5          & 63.8          & 28.3          & 32.5          & 11.1          & 7.5           & 25.0          & 31.1          & 12.2          \\
                                         & DAIMC\cite{DAIMC}                                        & 32.1          & 33.6          & 17.4          & \underline{57.5}    & \underline{78.7}    & 41.9          & 40.8          & 21.6          & 16.0          & 24.4          & 29.4          & 10.3          \\
                                         & EERIMVC\cite{EERIMVC}                                     & 39.6          & 39.0          & 22.1          & 49.0          & 74.2          & 34.2          & 33.2          & 14.3          & 3.9           & 24.9          & 29.6          & 12.2          \\
                                         & SURE\cite{yang2022SURE}                                         & 41.0          & 43.2          & 25.0          & 43.8          & 70.1          & 29.5          & 49.1          & 29.9          & 23.6          & 25.1          & 28.3          & 10.9          \\
                                         & DSIMVC\cite{DSIMVC}                                      & 31.7          & 35.6          & 17.2          & 19.7          & 40.0          & 19.7          & 43.2          & 23.3          & 19.0          & 18.1          & 18.6          & 5.6           \\
                                         & DCP\cite{DCP2022}                                  & 41.1          & \underline{45.1}    & 24.8          & 51.3          & 74.8          & \underline{51.9}    & 36.2          & 18.9          & 4.8           & \underline{26.2}    & \underline{32.7}    & \underline{13.5}    \\
                                         & ProImp\cite{ProImp2023}                                       & \underline{43.6}    & 45.0          & \underline{26.8}    & 37.6          & 67.0          & 25.0          & \underline{56.5}    & \underline{39.4}    & \underline{32.8}    & 23.7          & 27.9          & 10.8          \\
                                         & \textbf{DIVIDE(Ours)}                                         & \textbf{49.1} & \textbf{48.7} & \textbf{31.6} & \textbf{62.2} & \textbf{83.0} & \textbf{50.5} & \textbf{59.3} & \textbf{39.5} & \textbf{29.0} & \textbf{32.3} & \textbf{39.7} & \textbf{18.1} \\ \bottomrule
            \end{tabular}
    }
    \caption{The clustering performance on four multi-view benchmarks. The best and the second best results are denoted in bold and underline. ``Complete" setting denotes multi-view data without view missing and ``Incomplete" denotes 50\% samples are with missing views.}
    \label{table:main}
\end{table*}

\section{Experiments}

To evaluate the effectiveness of DIVIDE, we carry out extensive experiments to answer the following questions: 
(Q1) Does DIVIDE outperform state-of-the-art multi-view clustering methods?
(Q2) Could DIVIDE effectively handle multi-view clustering with missing views?
(Q3) Could the multi-step random walk effectively address the false negative issue?
(Q4) Does each component of DIVIDE contribute to the overall performance?

\subsection{Experimental Setup}
We briefly introduce the experimental setup here, including the used datasets, settings, and implementation details.

\subsubsection{Datasets.} We conduct evaluations on four widely-used multi-view benchmarks. In brief,
\begin{itemize}
    \item \textbf{Scene-15}~\citep{Scene15} contains 4,485 images of 15 scene categories. Following~\citet{yang2022SURE}, we use the PHOG and GIST features as two views.
    \item \textbf{Caltech-101}~\citep{Caltech} consists of 8,677 images of objects from 101 classes. Following~\citet{han2021trusted}, we utilize the deep features extracted by DECAF~\cite{krizhevsky2012imagenet} and VGG19~\citep{VGG19} as two views.
    \item \textbf{Reuters}~\cite{amini2009learning} is a multilingual news dataset with 18,758 samples from various languages. Following~\citet{MVSCN}, we project the English and French text into a 10-dim latent space using a standard autoencoder.
    \item \textbf{LandUse-21}~\cite{LandUse} contains 2,100 satellite images from 21 classes. Following~\citet{DCP2022}, we use PHOG and LBP features as two views.
\end{itemize}

\subsubsection{Experimental Setting.} Following most of the compared methods~\citep{DAIMC,EERIMVC,DCP2022,ProImp2023}, we obtain the clustering results by feeding the concatenated embedding of all views into $k$-means. For a comprehensive analysis, we use three widely used clustering metrics to evaluate the performance, including Normalized Mutual Information (NMI), Accuracy (ACC), and Adjusted Rand Index (ARI). A higher value of these metrics indicates a better clustering performance. Additionally, we verify the effectiveness of our method in incomplete MvC, \textit{i.e.}, some views of a sample might be missing during data collection. Specifically, we randomly select $m=\eta \times n$ samples and remove one view from each to simulate incomplete data, where $\eta$ is the missing rate and $n$ is the total number of samples. To handle missing views, we uses the cross-view decoders $g^{(v\rightarrow u)}$ to recover the representation of view $u$ via $\mathbf{z}_i^{(u)}=g^{(v\rightarrow u)}(\mathbf{z}_{k, i}^{(v)})$, where $v$ and $u$ denote the observed view and missing view, respectively.

\begin{figure*}[ht]
    \centering
    \includegraphics[width=0.325\textwidth]{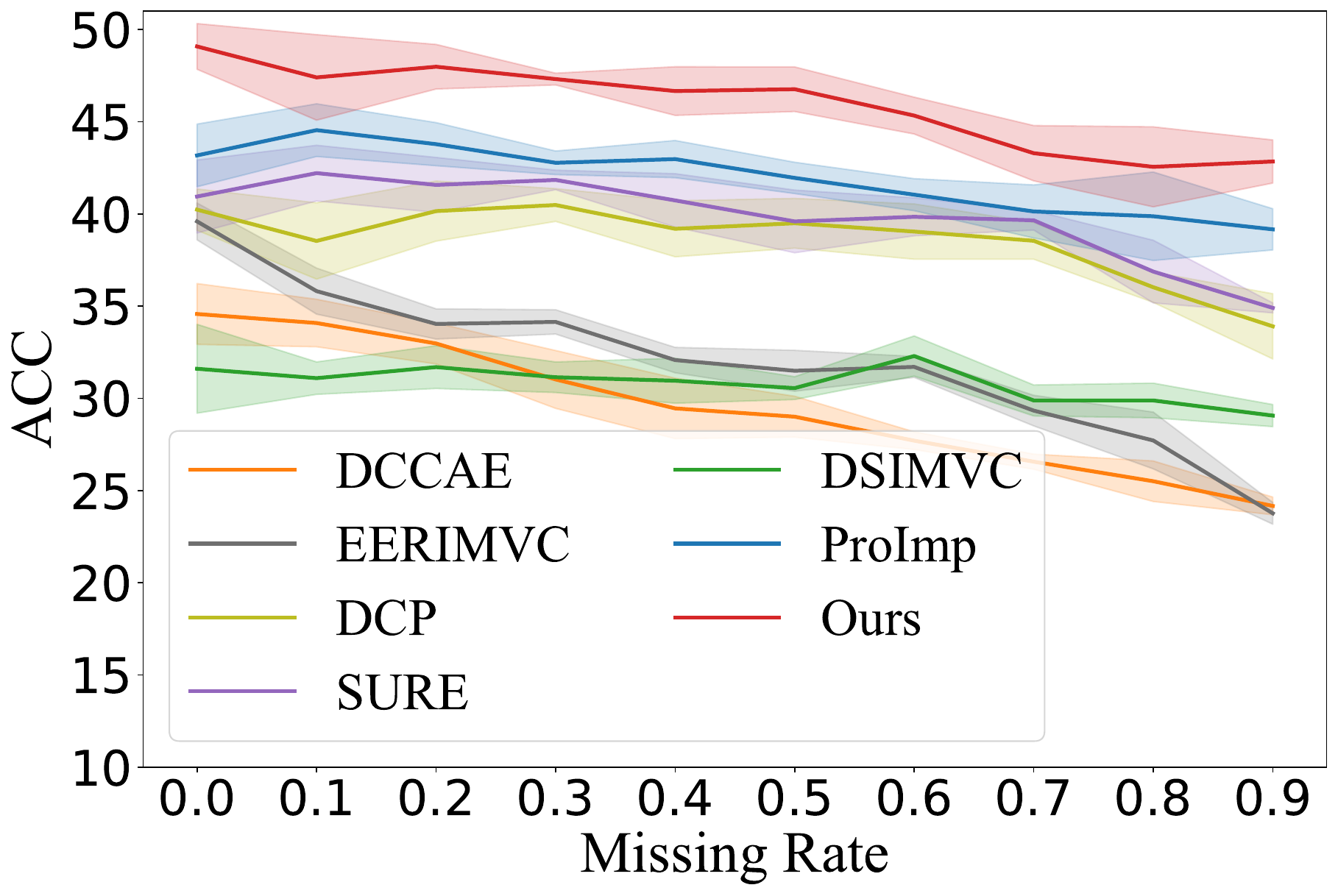}
    \hfill
    \includegraphics[width=0.325\textwidth]{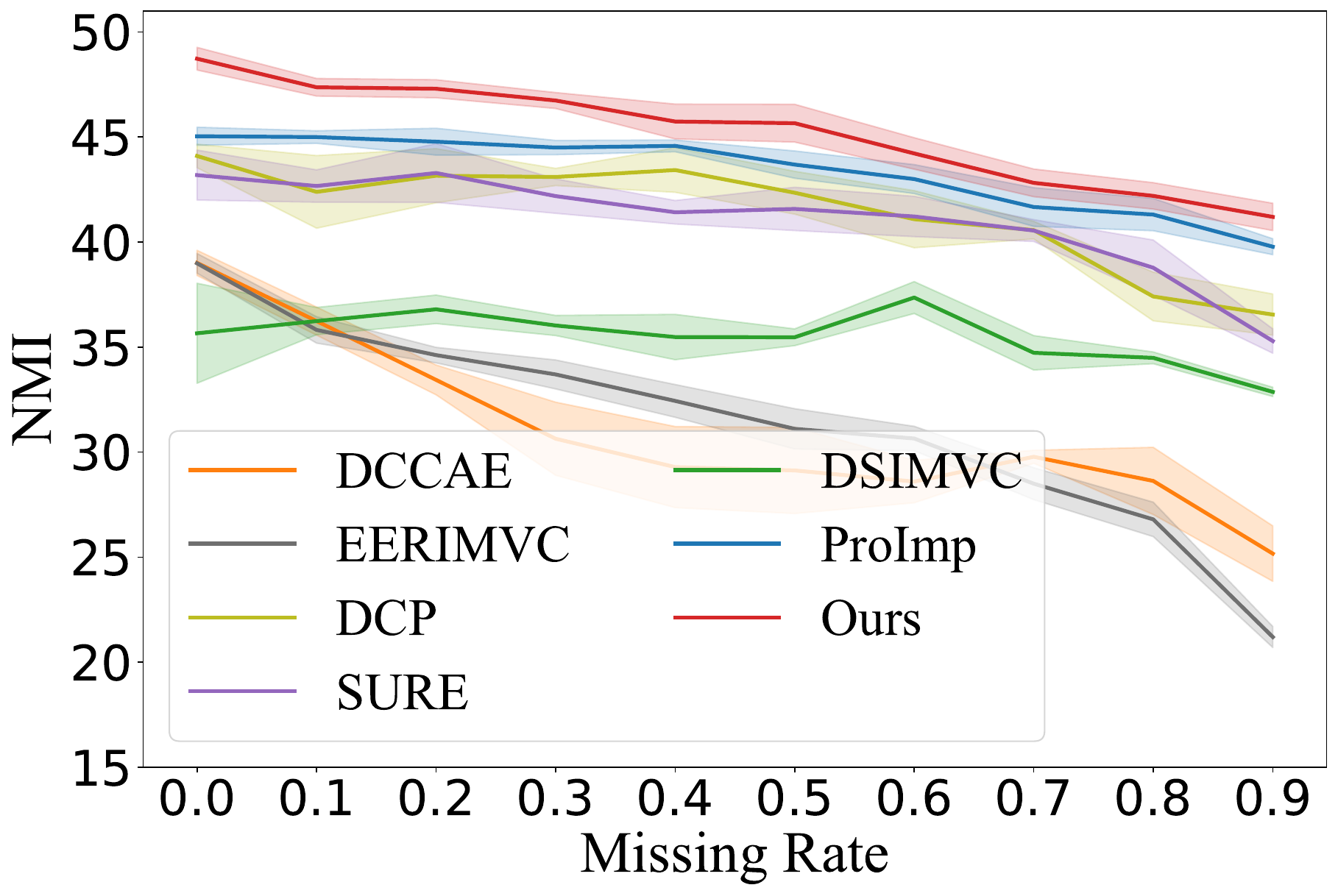}
    \hfill
    \includegraphics[width=0.325\textwidth]{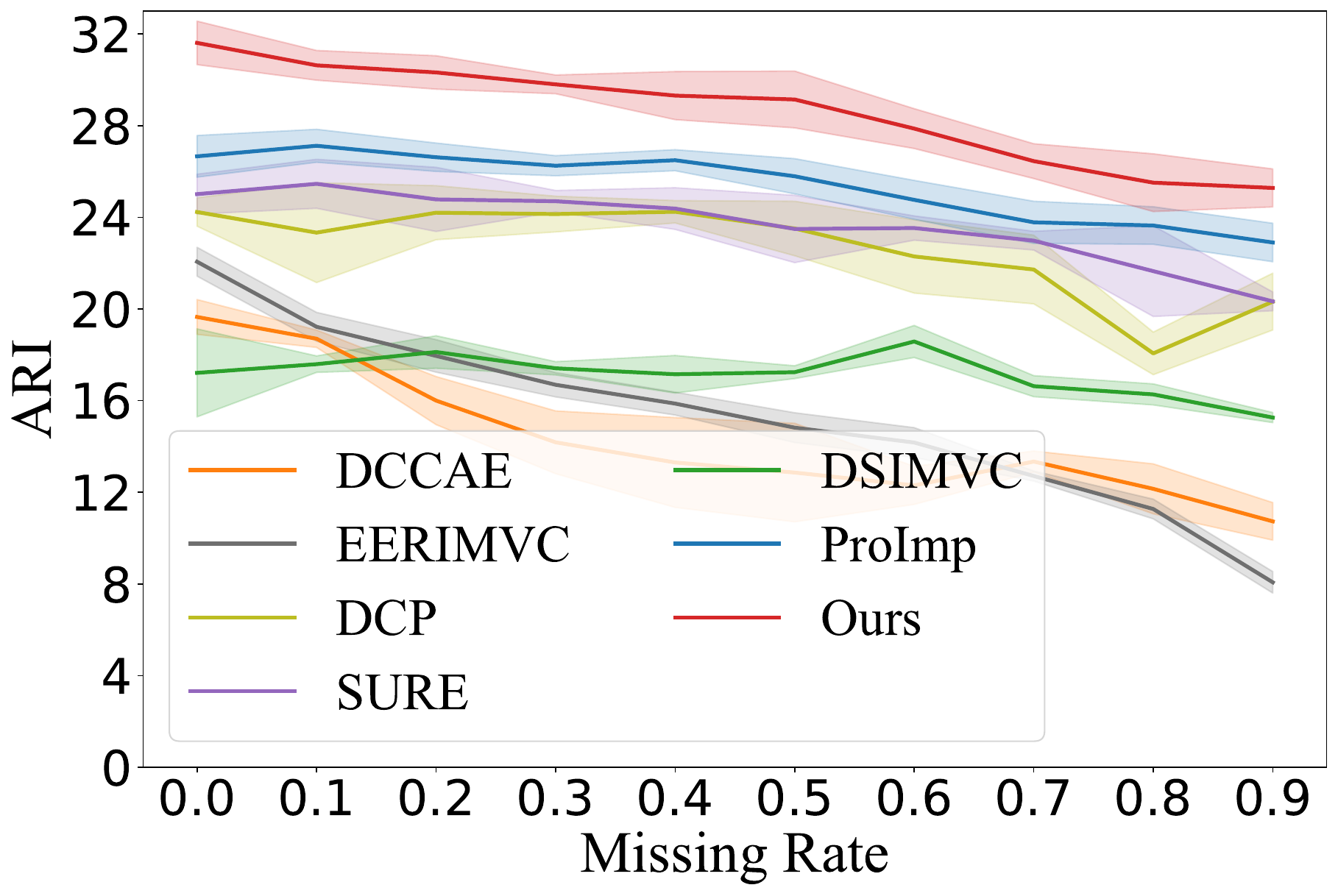}
    \caption{Clustering performance under different missing rates on Scene-15. The colored regions denote the standard variances with five random
    experiments.}
    \label{fig:missing_rate}
\end{figure*}

\subsubsection{Implementation Details.}
We implement our method in PyTorch 1.13.0 and run all experiments on NVIDIA 3090 GPUs in Ubuntu 20.04 OS. We train our model for 200 epochs using the Adam optimizer without weight decay. The initial learning rate is set to $2\times10^{-3}$, and the batch size is fixed to 1024.
To warm up, we set the target $\mathbf{T}$ in Eq.~(\ref{eq:loss}) as an identity matrix $\mathbf{I}_n$ for the first 100 epochs and then adopt the rectified target Eq.~(\ref{eq:t}) in the remaining epochs.
For the other hyper-parameters, we fix the contrastive temperature $\tau=0.5$, the temperature of the kernel function $\sigma=0.1$, the walking step $t=5$, and the rectified weight $\alpha=0.5$ throughout experiments. We employ a 4-layer fully connected network (FCN) with batch normalization and ReLU activation to build the view-specific encoder. The cross-view encoder is a 2-layer FCN and the details are placed in the supplementary material.

\subsection{Comparisons with State of the Arts (Q1)}

We compare our DIVIDE with nine state-of-the-art MvC methods, including DCCAE~\cite{DCCAE}, BMVC~\cite{BMVC}, PMVC~\cite{PMVC}, EERIMVC~\cite{EERIMVC}, DCP~\cite{DCP2022}, SURE~\cite{yang2022SURE}, DSIMVC~\cite{DSIMVC} and ProImp~\cite{ProImp2023}. As shown in Table~\ref{table:main}, DIVIDE achieves superior performance and greater robustness over the baselines in either complete or incomplete MvC settings. In particular, it significantly outperforms all baselines in the presence of missing views, indicating its capacity to handle incomplete MvC data. It is noteworthy that SURE~\cite{yang2022SURE} is also an FN-tolerant contrastive MvC method that uses a robust contrastive loss to reverse the gradient of FNs. The superiority of DIVIDE over SURE further demonstrates its robustness against FNs.

\subsection{Performance under Different Missing Rates (Q2)}

To investigate the robustness of our proposed method, we conduct experiments by changing the missing rates $\eta$. To be specific, we vary the missing rate $\eta$ from $0\%$ to $90\%$ with a gap of $10\%$ on the Scene-15 dataset. As the missing rate increases, the total number of samples might be smaller than the batch size. To avoid such a case, we set $\text{batch size}=\min(1024,(1-\eta)n)$. Fig.~\ref{fig:missing_rate} shows the experimental results, which indicate that DIVIDE outperforms all baselines in all incomplete evaluations.

\begin{table}[t]
    \tablestyle{2pt}{1.01}{
        \begin{tabular}{l|cc|cc}
\toprule
\multirow{2}{*}{Method}                                                                                      & \multicolumn{2}{c|}{Scene15} & \multicolumn{2}{c}{Caltech101} \\
& ACC           & KL & ACC            & KL\\ \midrule
Random Walk (Ours)  & \textbf{49.1} & \textbf{8.2} & \textbf{62.2}  & \textbf{5.1}  \\
$k$-nearest-neighbors~\cite{GCC2021}  & \underline{48.5}    & 22.3  & 57.0     & 17.2   \\
$\epsilon$-Neighborhood~\cite{yang2022SURE} & 48.4          & \underline{19.1}   & \underline{60.7}     & \underline{8.5}     \\
None                                                                                                         & 48.0            & 23.0           & 54.3           & 22.1          \\
\bottomrule
\end{tabular}%
}
    \caption{Clustering accuracy and KL divergence of different false negative identification strategies. ``None" denotes the case without FN identification. }
    \label{table:ACC_and_KL}
\end{table}

\subsection{Analysis on Multi-step Random Walks (Q3)}
In this section, we investigate the effectiveness of the proposed random walk strategy for FN identification and correction.

\begin{figure}[t]
    \centering
	\begin{minipage}[t]{0.495\columnwidth}
		\centering
		\includegraphics[width=\columnwidth]{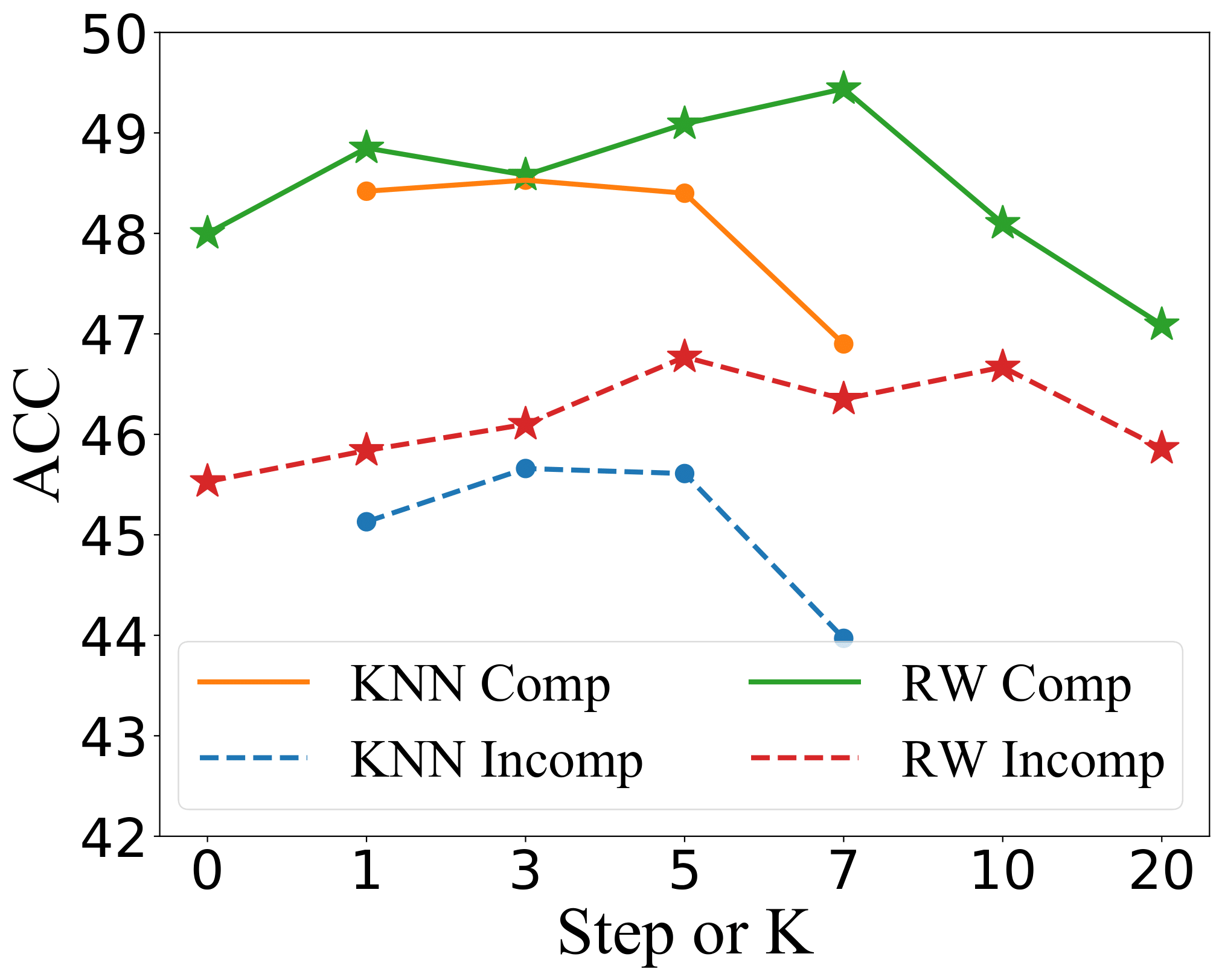}
        \subcaption{Scene-15}
	\end{minipage}
	\begin{minipage}[t]{0.495\columnwidth}
		\centering
		\includegraphics[width=\columnwidth]{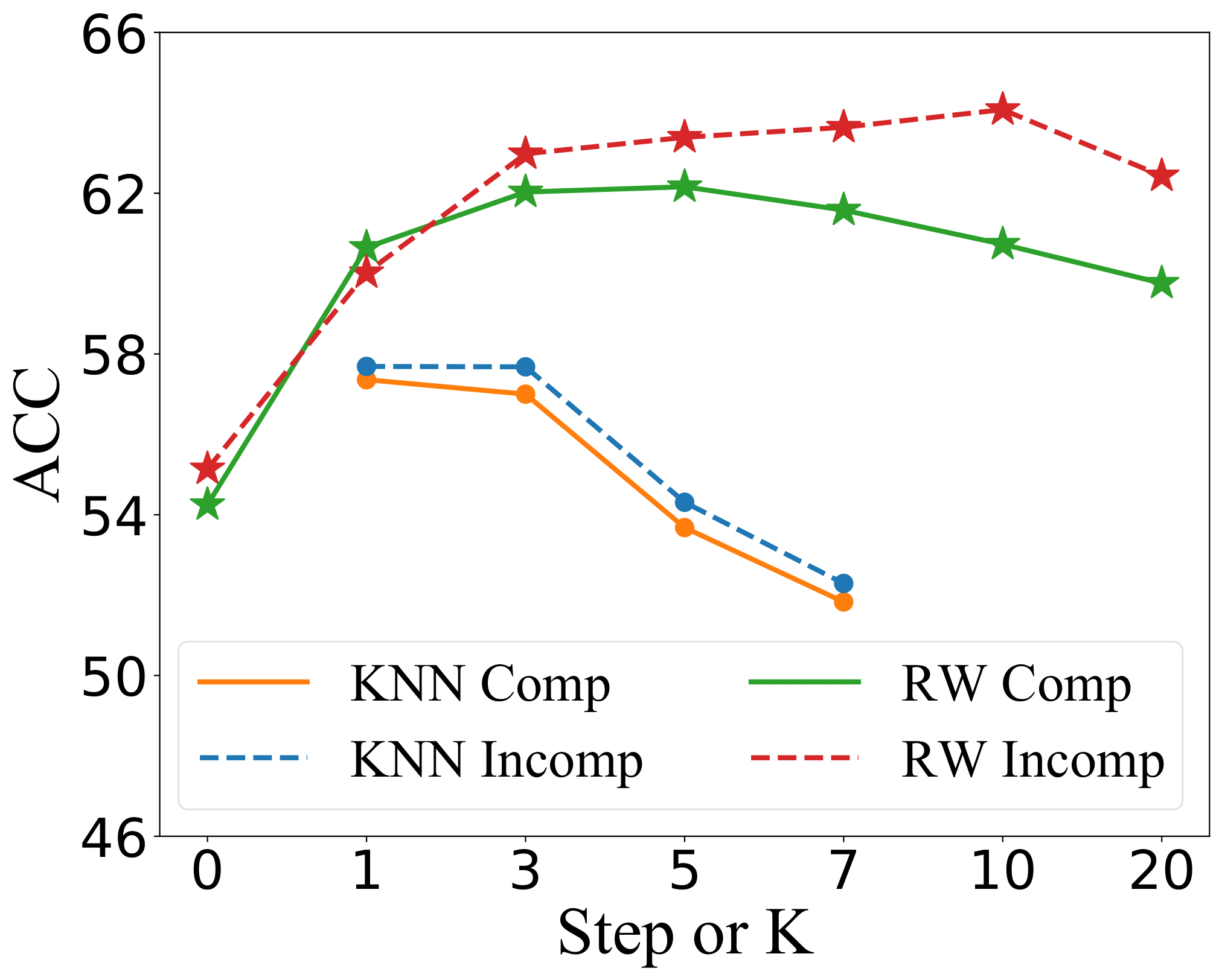}
		\subcaption{Caltech-101}
	\end{minipage}
    \caption{Clustering performance with different walking steps or different neighborhood sizes. ``KNN" denotes $k$-nearest-neighbors and ``RW" denotes our random walk paradigm. ``Comp" and ``Incomp" refer to complete and incomplete multi-view clustering settings.}
    \label{fig:step_ablation}
\end{figure}

\paragraph{Random Walk vs. $\epsilon$- and $k$-nearest-neighborhood.}
We compare our random walk strategy with the $\epsilon$- and $k$-nearest-neighbor-based FN identification strategies in terms of clustering accuracy and KL divergence.
The KL divergence is calculated between the rectified contrastive target $\mathbf{T}$ (Eq.~(\ref{eq:t})) and the oracle $\mathbf{T}^{\operatorname{GT}}$ constructed by the ground-truth labels. Specifically, $\mathbf{T}^{\operatorname{GT}}_{ij}= 1/|P_i|$ if the $i$-th and $j$-th samples belong to the same class, otherwise $0$. Here, $|P_i|$ is the number of in-batch samples that share the same label with $i$-th sample. 
We use the KL divergence to measure the performance of the rectification, \textit{i.e.}, $D_{\text{KL}}(\mathbf{T}^{\operatorname{GT}} \| \mathbf{T})$. 
Note that the KL divergence not only measures the accuracy of FN identification but also reflects the amount of incorrect rectified FP. A lower KL divergence score indicates more accurate correction, closer to the upper bound $\mathbf{T}^{\operatorname{GT}}$. As shown in Table ~\ref{table:ACC_and_KL}, our method remarkably improves the robustness, enabling more accurate rectification of FNs while avoiding introducing FPs. Besides, we compare our random walk strategy with the $k$-nearest-neighbor strategy by changing walking steps $t$ for the former and neighborhood size $k$ for the latter. Note that, $t=0$ corresponds to the case without rectification, \textit{i.e}. $\mathbf{T}^0=\mathbf{I}_n$. Fig.~\ref{fig:step_ablation} shows that the higher-order random walk achieves better performance than the lower ones, verifying the effectiveness of our method. In addition, one could observe that DIVIDE is insensitive to the walk step $t$, which achieves encouraging results with 3--7 steps. In contrast, the $k$-nearest-neighbor strategy will dramatically drop the performance if $k$ is not elaborately selected. 

\begin{table}[t]
\tablestyle{8pt}{1.01}{
\begin{tabular}{c|ccc}
\toprule
\multirow{2}{*}{Strategy} & \multicolumn{3}{c}{Scene-15}                  \\
                                  & ACC           & NMI           & ACC           \\ \midrule
self \& swap                      & \textbf{49.1} & \textbf{48.7} & \textbf{31.6} \\
self                              & 48.7          & 48.5          & 31.2          \\
swap                              & 48.7          & 47.9          & 30.8          \\
concat                            & 48.1          & 48.1          & 30.5          \\ \bottomrule

\end{tabular}
}
\caption{Influence of different targets for Eq.~(\ref{eq:loss}).
``self \& swap" is the default strategy as discussed in Method.
``self" denotes that the pseudo target $\mathbf{T}^{(v)}$ is obtained from the affinity graph of the given view $v$, ``swap" denotes that the target is obtained from the graph of the predicted view $u$. ``concat" means that the target is obtained from the fused affinity graph constructed by concatenating all views' embedding.
}
\label{table:compare_strategy}
\end{table}

\paragraph{Different Choices of the Target $\mathbf{T}^{(v)}$.} {We perform experiments by adopting different rectified targets $\mathbf{T}^{(v)}$ for the decoupled contrastive loss (Eq.~(\ref{eq:loss})). 
As shown in Table~\ref{table:compare_strategy}, DIVIDE consistently performs well with various rectified targets, demonstrating the effectiveness of our random-walk-based false negative (FN) identification module. 
Specifically, the ``self \& swap" strategy achieves better results, which enjoys the merit of both the self target and the swap target, where the self target preserves the intra-view information better and the swap target facilitates the cross-view interactions. }

\begin{table}[t]
    \tablestyle{5pt}{1.01}{
        \begin{tabular}{cccc|ccc}
\toprule
\multirow{2}{*}{$\mathcal{L}_{\text{inter}}$} &
  \multirow{2}{*}{$\mathcal{L}_{\text{intra}}$} &
  \multirow{2}{*}{decoder} &
  \multirow{2}{*}{rectify} &
  \multicolumn{3}{c}{Scene-15} \\
             &              &              &              & ACC        & NMI        & \multicolumn{1}{l}{ACC} \\ \midrule
             & $\checkmark$ &              &              & 39.6       & 39.7       & 22.5                    \\
    $\checkmark$ &     &     &     & 43.7       & 43.6       & 26.1                    \\
$\checkmark$ & $\checkmark$ &              &              & 47.4       & 46.5       & 29.6                    \\
$\checkmark$ & $\checkmark$ &              & $\checkmark$ & \underline{48.4} & 47.3       & 30.4                    \\
$\checkmark$ & $\checkmark$ & $\checkmark$ &              & 48.0       & \underline{48.1} & \underline{31.0}              \\
$\checkmark$ &
  $\checkmark$ &
  $\checkmark$ &
  $\checkmark$ &
  \textbf{49.1} &
  \textbf{48.7} &
  \textbf{31.6} \\ \bottomrule

\end{tabular}%
    }
    \caption{Ablation study of our method on Scene-15, where $\checkmark$ denotes the components is adopted..}
    \label{table:ablation}
\end{table}

\begin{table}[t]
    \tablestyle{4pt}{1.01}{
    \begin{tabular}{ccc|ccc}
\toprule
\multirow{2}{*}{decoder} & \multirow{2}{*}{target encoder} & \multirow{2}{*}{stop gradient} & \multicolumn{3}{c}{Scene-15}                  \\
             &          &              & ACC         & NMI           & ARI           \\ \midrule
             & share    &              & 44.0          & 44.3          & 26.7 \\
             & share    & $\checkmark$ & 46.7        & 46.3          & 29.7          \\
             & momentum & $\checkmark$ & \underline{48.4}  & \underline{47.3}    & \underline{30.4}    \\
$\checkmark$ & share    &              & 43.1        & 44.1          & 25.9          \\
$\checkmark$ & share    & $\checkmark$ & 47.6        & 45.7          & 28.2          \\
$\checkmark$             & momentum                        & $\checkmark$                   & \textbf{49.1} & \textbf{48.7} & \textbf{31.6} \\ \bottomrule
\end{tabular}%
    }
    \caption{Ablation study of momentum network and cross-view decoder.}
    \label{table:ablation_stop_gradient}
\end{table}

\subsection{Ablation Studies (Q4)}
In this section, we evaluate the role of different components of DIVIDE on the Scene-15 dataset. 

\paragraph{Ablation on Model Components.}
We first investigate the influence of different DIVIDE components and further explore some variations of the decoupled contrastive learning paradigm. 
As shown in Table.~\ref{table:ablation}, we isolate the effects of the intra-view contrastive loss ($\mathcal{L}_{\text{intra}}$), the inter-view contrastive loss ($\mathcal{L}_{\text{inter}}$), the cross-view decoder, and the false negative (FN) rectification strategy. When the cross-view decoder is removed, we directly apply $\mathcal{L}_{\text{inter}}$ to $\mathbf{z}_q$ and $\mathbf{z}_k$. The experimental results show that  $\mathcal{L}_{\text{intra}}$ could enhance the intra-view complementarity, leading to $3.7\%$ and $2.9\%$ improvement in terms of ACC and NMI, respectively. The cross-view decoder helps the model achieve a $0.6\%$ improvement in ACC and a $1.6\%$ improvement in NMI. Moreover, the rectified contrastive objective achieves a gain of $1\%$ in ACC in all settings with and without the cross-view decoder. The results suggest that all components of DIVIDE contribute to the state-of-the-art performance in MvC tasks.

\paragraph{Ablation of Decoupled Contrastive Learning.} 
We conduct further exploration of the decoupled contrastive learning framework by investigating the effectiveness of the momentum updating strategy and the stop-gradient operation of the target encoders. In the experiments, ``share'' refers to directly using the online encoder as the target encoder. The stop-gradient operation refers to stopping the gradient flow of the target encoder. As shown in Table~\ref{table:ablation_stop_gradient}, the performance of is significantly affected when either of these components is removed, indicating that the momentum updating strategy and stop-gradient operation play crucial roles in preserving view-specific information. 

\section{Conclusion}
This paper proposes a robust contrastive multi-view clustering method that improves existing works in two aspects: 
i) achieving robustness against missing views through using a decouple contrastive learning framework. In addition, the framework could get an elegant balance between view-specific information preservation and cross-view consistency learning, thus facilitating the MvC performance; 
ii) achieving robustness against false negatives through using a multi-step random walk to identify the false negatives. 
As a result, both in-neighborhood negatives and out-of-neighborhood positives could be correctly identified, thus alleviating the false positive and false negative issues.
Extensive experiments verify the effectiveness of the proposed method in both complete and incomplete MvC scenarios. 
In the future, we would like to explore how to extend our method to improve the robustness of contrastive representation learning by specifically incorporating its characteristics. 

\section*{Acknowledgments}
This work was supported in part by NSFC under Grant U21B2040, 62176171, and 62102274; in part by the Fundamental Research Funds for the Central Universities under Grant CJ202303.

\section{Supplementary Material}
In supplementary material, we provide experiment details about the network architectures, training hyper-parameters, and more visualization results of our method. 

\subsection{Training Implementation}
The proposed decouple contrastive learning framework contains two learnable modules, \textit{i.e.} the view-specific Siamese encoder and cross-view decoder. We employ the Fully Connected Network (FCN) for both two modules, where each layer is sequentially followed by a Batch Normalization~\cite{ioffe2015batch} and a Rectified Linear Unit (ReLU) activation function. We introduce an additional Batch Normalization layer as the final layer of the encoder, as recommended in~\citet{MoCo_v3}, to enhance training stability. 
For the view-specific Siamese encoder, the structure of the target encoder $f_k^{(v)}$ mirrors that of the online encoder $f_q^{(v)}$. The parameters of the target encoder are updated by Exponential Moving Average (EMA) of the online encoder. Formally, denoting the parameters of the online and target encoder as $\theta_q^{(v)}$ and $\theta_k^{(v)}$, respectively. The target encoder is updated by,
\begin{equation}
    \theta_k^{(v)}\leftarrow m\theta_k^{(v)}+(1-m)\theta_q^{(v)},
\end{equation}
where $m\in(0,1]$ is a fixed momentum coefficient during training. We set $m=0.98$ in our experiments.
The details of the network architectures has presented in Table~\ref{table:network_arch} and the training hyper-parameters is placed in Table~\ref{label:training_configuration}.

\subsection{Visualization of Multi-step Random Walk}
\begin{figure*}[h]
    \centering
	\begin{minipage}[t]{0.85\textwidth}
		\centering
		\includegraphics[width=\textwidth]{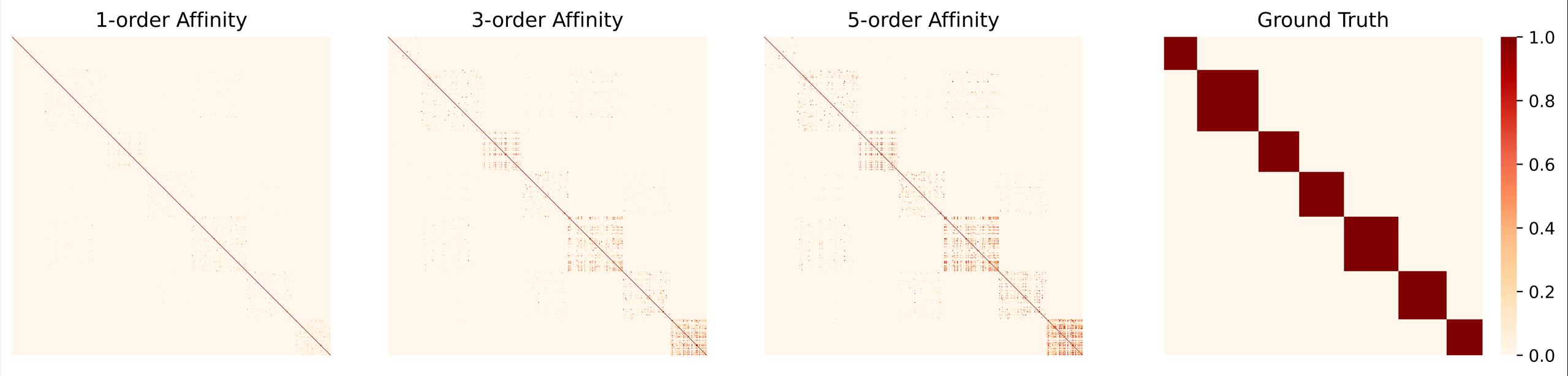}
        \subcaption{Scene-15}
	\end{minipage}\\

	\begin{minipage}[t]{0.85\textwidth}
		\centering
		\includegraphics[width=\textwidth]{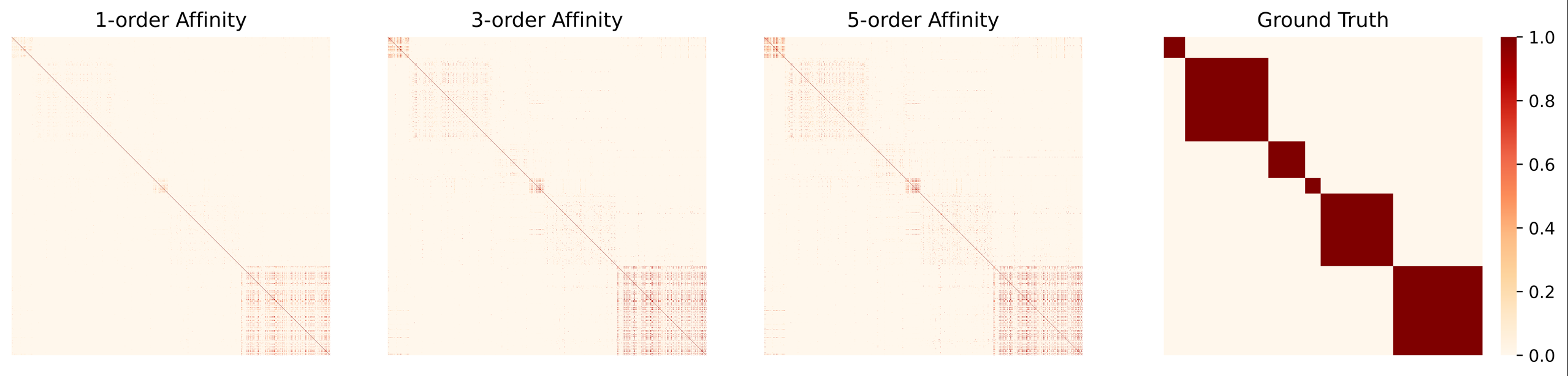}
		\subcaption{Reuters}
	\end{minipage}
    \caption{Transition Matrix of Scene15 and Reuters.}
    \label{fig:high_order_affinity}
\end{figure*}

In this section, we investigate the transition matrix obtained by the multi-step random walk to show the effectiveness of the false negative identification. 
We visually juxtapose the transition matrix $\mathbf{M}^t,\ t\in\{1,3,5\}$ alongside the corresponding ground truth matrix, which is constructed based on the class labels, in Fig.~\ref{fig:high_order_affinity}. As depicted, as the number of random walk steps increases, the transition matrix exhibits an increasingly accurate ability to identify false negatives. This compelling observation proves the effectiveness of our method to discover more intra-class samples.

\begin{table}[t]
    \tablestyle{15pt}{1.1}{
    \begin{tabular}{|c|c|}
    \hline
    Configure               & Value            \\ \hline
    Optimizer            & Adam             \\
    Batch size           & 1024             \\
    Learning rate        & $2\times10^{-3}$ \\
    Total training epoch                & 200              \\
    Start rectify epoch  & 100              \\ 
    Momentum coefficient  & 0.98              \\ \hline

    \end{tabular}%
    }
    \caption{Training setup.}
    \label{label:training_configuration}
\end{table}

\subsection{Visualization of Learned Representation}
\begin{figure*}[h]
    \centering
    \begin{minipage}[t]{0.2\textwidth}
	\centering
	\includegraphics[width=\textwidth]{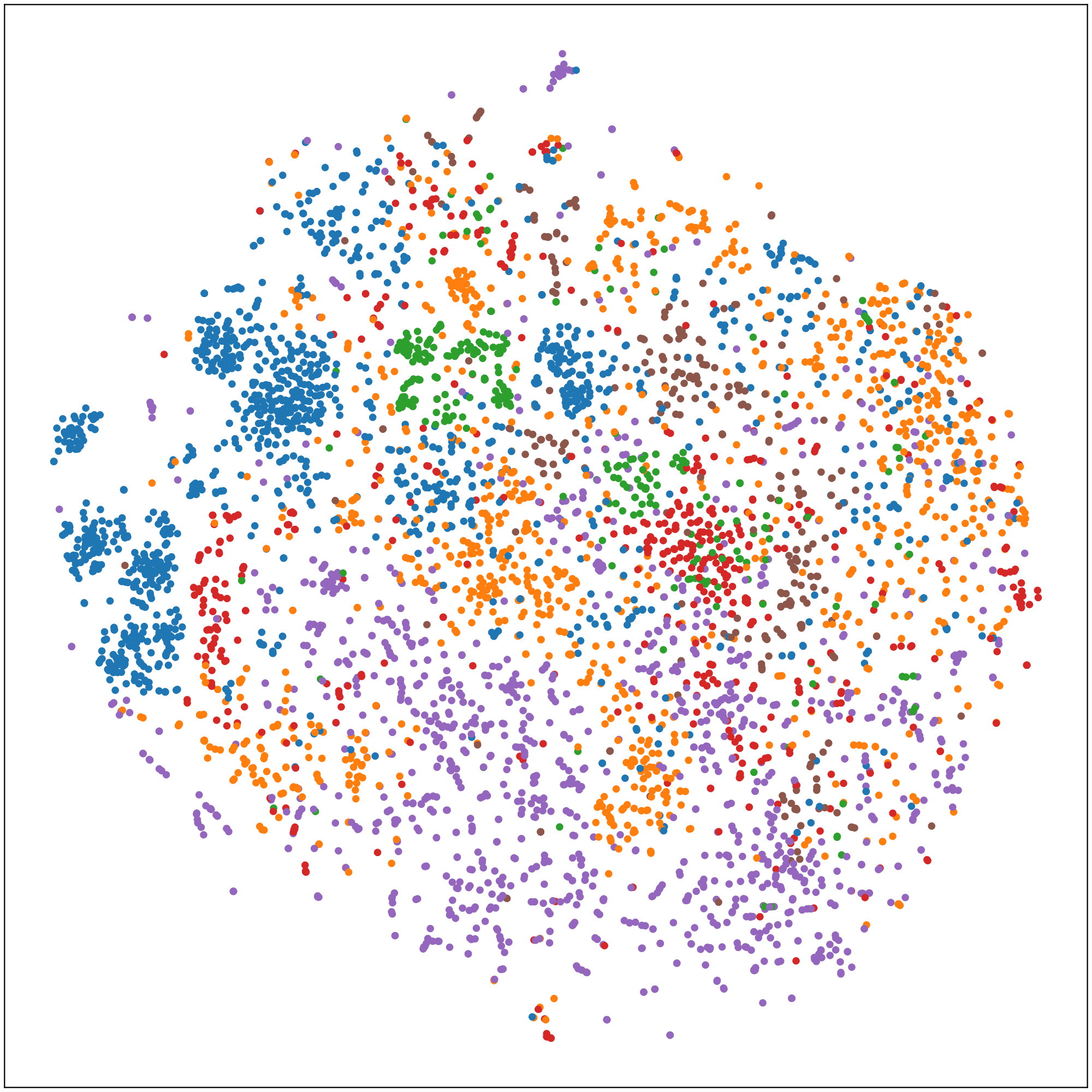}
        \subcaption{Epoch 0\\(ACC=36.5\%)}
	\end{minipage}
    \hspace{0.05in}
    \begin{minipage}[t]{0.2\textwidth}
	\centering
	\includegraphics[width=\textwidth]{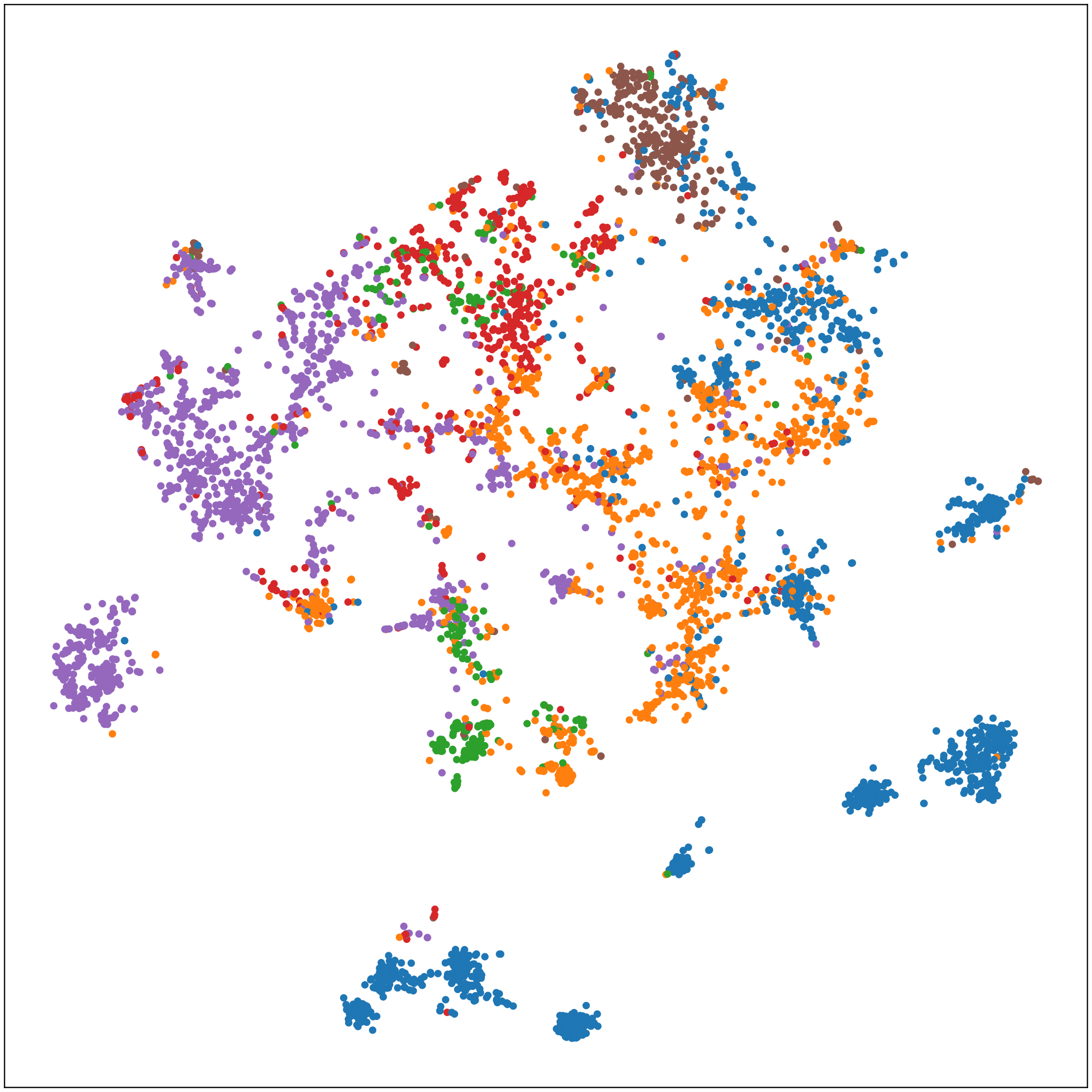}
        \subcaption{Epoch 100\\(ACC=53.9\%)}
	\end{minipage}
    \hspace{0.05in}
    \begin{minipage}[t]{0.2\textwidth}
	\centering
	\includegraphics[width=\textwidth]{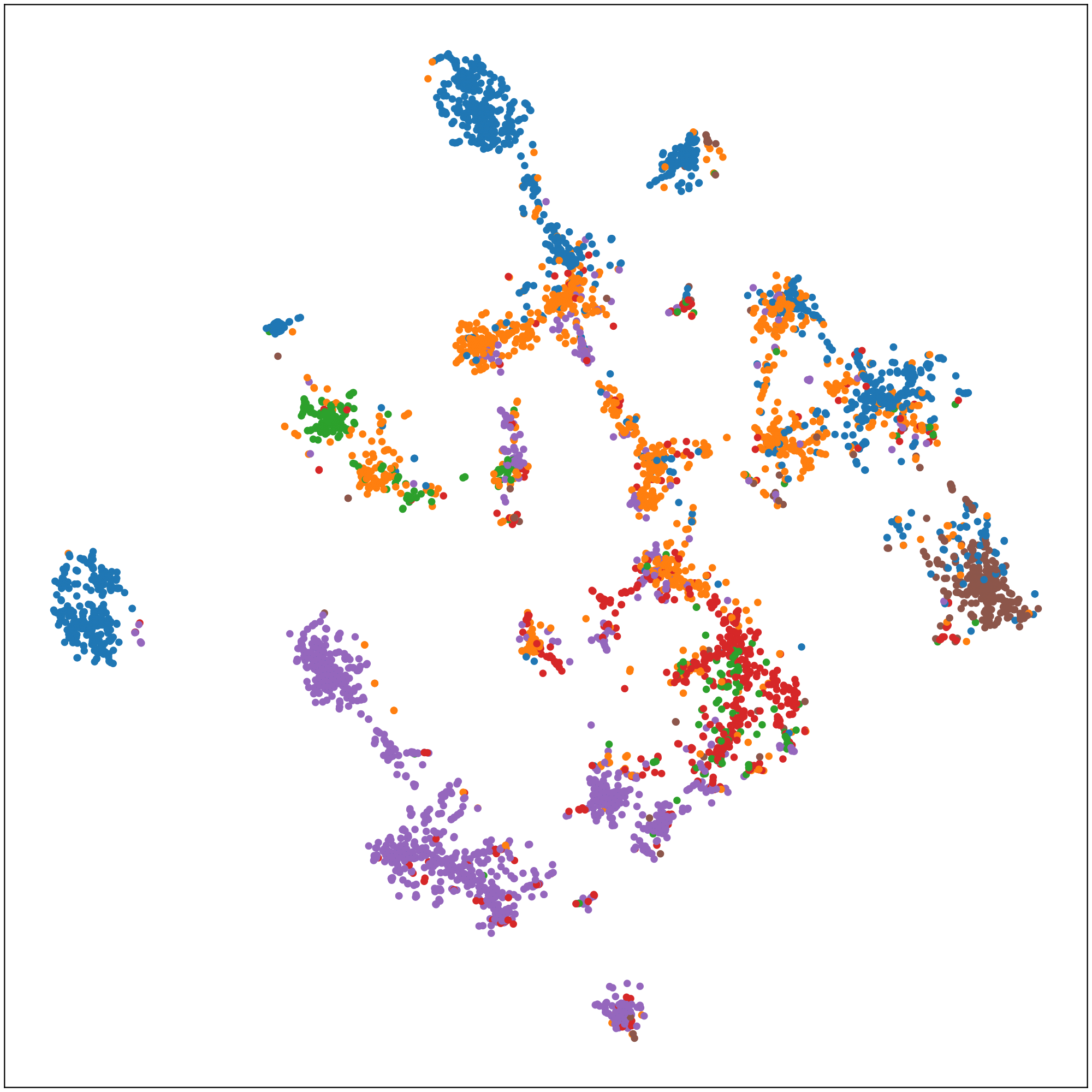}
        \subcaption{Epoch 200 with RW\\(ACC=59.1\%)}
        \label{fig:learning_representation_c}
	\end{minipage}
    \hspace{0.05in}
    \begin{minipage}[t]{0.2\textwidth}
	\centering
	\scalebox{-1}[1]{\includegraphics[width=\textwidth]{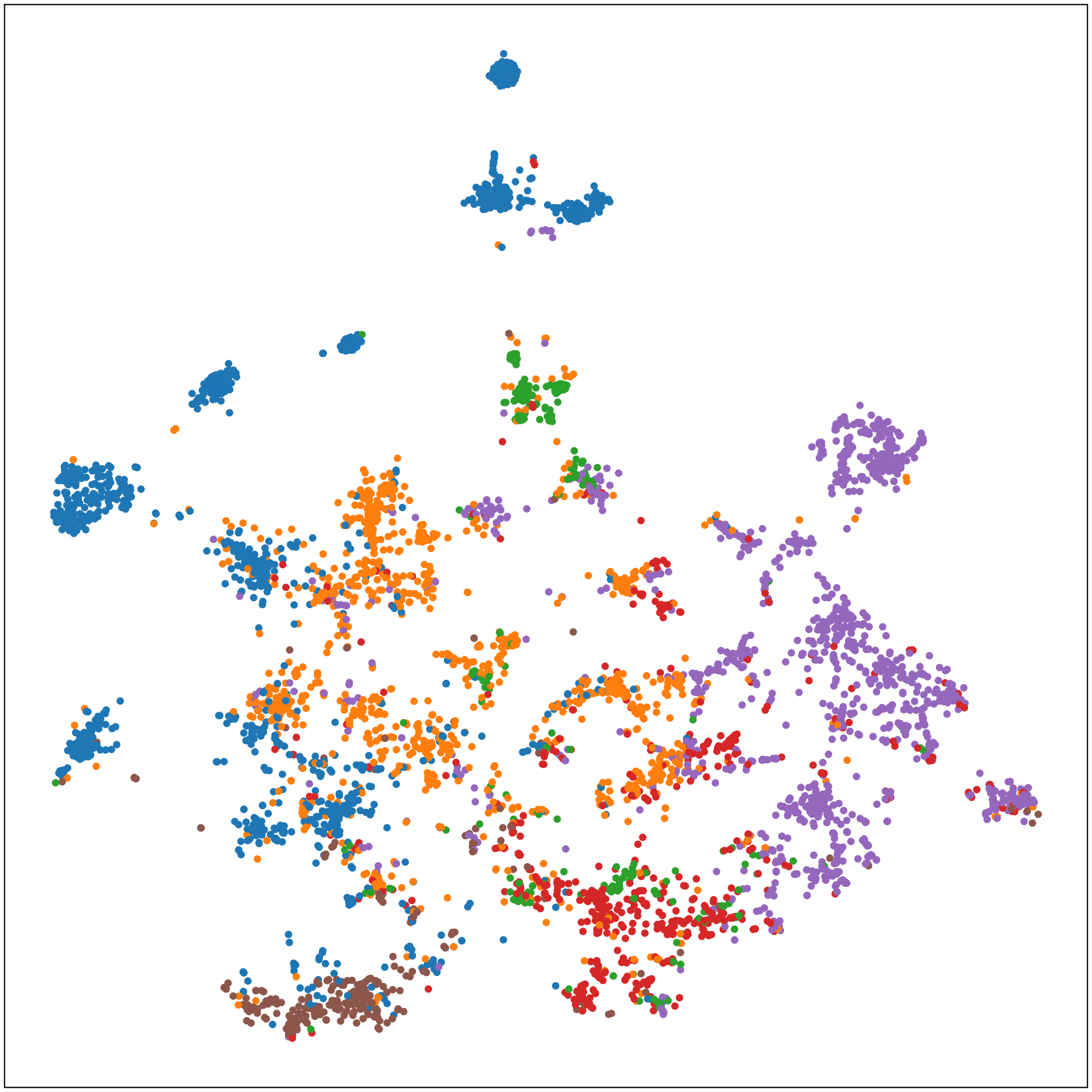}}
        \subcaption{Epoch 200 without RW\\(ACC=56.2\%)}
        \label{fig:learning_representation_d}
	\end{minipage}
    \caption{The t-SNE visualization on the Reuters dataset across the training process. After the 100 epoch, we apply the multi-step random walk to identify the false negative. ``RW'' denotes the random-walk-based identification.}
    \label{fig:learning_representation}
\end{figure*}

In this section, we further present the visualization of learned representation by our method. 
Specifically, we employ t-SNE~\cite{tsne}, a widely used technique for visualizing high-dimensional data, to visualize the features of the Reuters dataset across various training epochs. 
As shown in Fig.~\ref{fig:learning_representation}, we observe the initial state of the representation appears rather randomly dispersed. 
As the training progresses to the 100 epoch, the intra-cluster samples tend to form multiple sub-clusters that are widely separated from each other. This phenomenon is especially observed in the case of the blue class. 
However, upon completion of the training process (\textit{i.e.,} 200 epoch), the clusters become more compact, as can be seen in Fig.~\ref{fig:learning_representation_c}. This improvement in cluster compactness is not observed in the case of DIVIDE without false negative identification (Fig.~\ref{fig:learning_representation_d}). 
The visual analysis of the learning representation clearly demonstrates the effectiveness of our random-walk-based false negative identification approach.

\begin{table*}[t]
    \centering
    \begin{minipage}[t]{0.75\textwidth}
    \tablestyle{12pt}{1.01}{
        \begin{tabular}{|c|c|c|}
            \hline
            Dataset &
              View-specific Encoder &
              Cross-view Decoder \\ \hline
            Scene-15 &
              \begin{tabular}[c]{@{}c@{}}Linear($\text{dim}^{(v)}$,~1024),~BatchNorm,~ReLU\\      Linear(1024,~1024),~BatchNorm,~ReLU\\      Linear(1024,~ 1024),~BatchNorm,~ReLU\\      Linear(1024,~128),~BatchNorm\end{tabular} &
              \begin{tabular}[c]{@{}c@{}}Linear(128,~512), ReLU\\      Linear(512,~128)\end{tabular} \\ \hline
            Caltech101 &
              \begin{tabular}[c]{@{}c@{}}Linear($\text{dim}^{(v)}$,~1024),~BatchNorm,~ ReLU\\      Linear(1024,~1024),~BatchNorm,~ReLU\\      Linear(1024,~1024),~BatchNorm,~ReLU\\      Linear(1024,~128),~BatchNorm\end{tabular} &
              \begin{tabular}[c]{@{}c@{}}Linear(128,~512),~ReLU\\      Linear(512,~128)\end{tabular} \\ \hline
            Reuters &
              \begin{tabular}[c]{@{}c@{}}Linear($\text{dim}^{(v)}$,~128),~BatchNorm,~ReLU\\      Linear(128,~1024),~BatchNorm,~ReLU\\      Linear(1024,~1024),~BatchNorm,~ReLU\\      Linear(1024,~64),~BatchNorm\end{tabular} &
              \begin{tabular}[c]{@{}c@{}}Linear(64,~256),~ReLU\\      Linear(256,~64)\end{tabular} \\ \hline
            LandUse21 &
              \begin{tabular}[c]{@{}c@{}}Linear($\text{dim}^{(v)}$,~1024),~BatchNorm,~ReLU\\      Linear(1024,~1024),~BatchNorm,~ReLU\\      Linear(1024,~1024),~BatchNorm,~ReLU\\      Linear(1024,~128),~BatchNorm\end{tabular} &
              \begin{tabular}[c]{@{}c@{}}Linear(128,~512),~ReLU\\      Linear(512,~128)\end{tabular} \\ \hline
            \end{tabular}%
    }
    \end{minipage}
    \caption{The architecture of the view-specific encoders and cross-view decoders in DIVIDE. ``$\text{dim}^{(v)}$'' denotes the dimension of input data of the $v$-th view.}
    \label{table:network_arch}
\end{table*}

\bibliography{aaai24}

\end{document}